\documentclass[sigconf, nonacm]{custom-arxiv}

\usepackage{balance}
\usepackage{float}
\usepackage{graphicx}
\usepackage{caption}
\usepackage{subcaption}

\title[FCMNet: Full Communication Memory Net]{FCMNet: Full Communication Memory Net for Team-Level Cooperation in Multi-Agent Systems}

\author{Yutong Wang}
\affiliation{
  \institution{National University of Singapore \\ Department of Mechanical Engineering}
  \city{}
  \country{Republic of Singapore}}
\email{e0576114@u.nus.edu}

\author{Guillaume Sartoretti}
\affiliation{
  \institution{National University of Singapore \\ Department of Mechanical Engineering}
  \city{}
  \country{Republic of Singapore}}
\email{mpegas@nus.edu.sg}

\begin{abstract}

Decentralized cooperation in partially-observable multi-agent systems requires effective communications among agents.
To support this effort, this work focuses on the class of problems where global communications are available but may be unreliable, thus precluding differentiable communication learning methods.
We introduce FCMNet, a reinforcement learning based approach that allows agents to simultaneously learn a) an effective multi-hop communications protocol and b) a common, decentralized policy that enables team-level decision-making.
Specifically, our proposed method utilizes the hidden states of multiple directional recurrent neural networks as communication messages among agents.
Using a simple multi-hop topology, we endow each agent with the ability to receive information sequentially encoded by every other agent at each time step, leading to improved global cooperation.
We demonstrate FCMNet on a challenging set of StarCraft II micromanagement tasks with shared rewards, as well as a collaborative multi-agent pathfinding task with individual rewards.
There, our comparison results show that FCMNet outperforms state-of-the-art communication-based reinforcement learning methods in all StarCraft II micromanagement tasks, and value decomposition methods in certain tasks.
We further investigate the robustness of FCMNet under realistic communication disturbances, such as random message loss or binarized messages (i.e., non-differentiable communication channels), to showcase FMCNet’s potential applicability to robotic tasks under a variety of real-world conditions.

\end{abstract}


\begin{document}

\pagestyle{fancy}
\fancyhead{}

\maketitle


\section{Introduction}

Multi-Agent Reinforcement Learning (MARL) has attracted a lot of attention~\cite{kiran2021deep,dafoe2020open} in recent years, supported by the advances in single-agent Reinforcement Learning and the development of advanced neural structures.
It is showing promises for a wide array of real-life applications, such as motion planning for autonomous vehicle~\cite{kiran2021deep,wang2021autonomous}, multi-robot control~\cite{damani2021primal,sartoretti2019distributed}, in addition to applications to video games AI~\cite{arulkumaran2019alphastar,berner2019dota}.
However, the switch from single- to multi-agent RL brings distinct new challenges.
In particular, MARL agents encounter partially-observable environments in many real-life tasks, where learning effective cooperative policies solely based on their own knowledge/memory can be challenging or outright impossible.
Partial observability of the world is also made worse by the fact that the intention of other agents is often unknown/unmodelled, limiting agents to seeing each other as dynamic obstacles in the world.
One solution to these issues is the introduction of explicit communication among agents, to share relevant information and/or intents, towards true joint cooperation at the team-level.
That is, communication enables decentralized agents to behave as a group, rather than as a collection of individuals.

\vfill

This work focuses on \textit{communication learning} (CL), where agents are tasked with simultaneously learning a communication protocol, to identify, encode, and share relevant information, as well as a cooperative action policy conditioned upon received information.
Early work in the field proposed two ways to learn a communication policy, allowing agents to select what information to send each other at each time step~\cite{foerster2016learning}.
RIAL trains a discrete policy (i.e., choosing one among a pre-defined set of messages) using standard reinforcement learning (RL) from individual rewards (i.e., reinforced CL), while DIAL trains a continuous-valued policy via backpropagation through the \textit{communication channel} that connects the agents (i.e., differentiable CL).
More recent works, such as SchedNet~\cite{kim2019learning}, G2ANet~\cite{liu2020multi}, ATOC~\cite{jiang2018learning}, have focused on more general communication learning, often relying on differentiable CL.
In these methods, the main focus is on allowing agents to 1) send/utilize multiple independent messages under dynamic communication topologies, and/or 2) select whom to communicate with and at which time steps, to reduce the overall communication burden in large teams.
However, as a result, these works do not usually make use of information from all agents, thus limiting team-wide cooperation, and often lack robustness investigations (e.g., resilience to message loss), which may limit their applicability under real-life conditions.

\vfill

In this paper, we focus on the class of problems where global communications are available but may be unreliable and introduce a new differentiable CL framework called Full Communication Memory Net (FCMNet).
Our method allows agents to simultaneously learn a global multi-hop communications protocol and a common, decentralized policy for cooperative tasks.
To this end, FCMNet utilizes the hidden states and cell states of multiple parallel directional recurrent neural networks as communication messages among agents.
At every timestep, each agent receives multiple messages from all other agents as well as information about its past observations (self-memory).
By relying on our proposed neural structure, and on weight sharing among agents in careful ways throughout our framework, FCMNet can be trained using differentiable CL and exhibits faster, more stable training to higher-cooperation policies than existing CL methods.

\vfill

We experimentally evaluate our proposed model on a range of unit micromanagement tasks in the StarCraft II Multi-Agent Challenge~\cite{samvelyan2019starcraft}, as well as on a partially-observable multi-agent
pathfinding \hfill task~\cite{freed2020communication}. \hfill
Our \hfill results \hfill show \hfill that \hfill FCMNet \hfill outperforms
\newpage \noindent
existing state-of-the-art CL methods in all StarCraft II micromanagement tasks and value decomposition methods in certain tasks.
We investigate the robustness of FCMNet under realistic communication interference, such as binary messages (digital communication, thus non-differentiable CL), random message loss, and randomized communication orders at each timestep.
Our findings show that FCMNet exhibits a form of natural resilience to such communication interference, showing promises for deployments in real-life tasks such as multi-robot deployments.


\section{Related work}

Foerster et al.\cite{foerster2016learning} first proposed two learnable communication protocols based on deep Q-networks, called RIAL (\textit{Reinforced Inter-Agent Learning}) and DIAL (\textit{Differentiable Inter-Agent Learning}), thus coining these two classes of approaches to communication learning (CL).
RIAL treats communication as an action to be selected, while DIAL learns real valued messages which are discretized at execution time.
As a result of handling real-valued messages, DIAL can directly push gradients from one agent to another through the (differentiable, noise-free) communication channel, which brings richer feedback to the agents to train a more effective communication channel.
More recently, CommNet~\cite{sukhbaatar2016learning} focused on global communication for fully cooperative tasks.
There, all agents are controlled by a single network, where communication channels are built to transmit the average hidden state of all other agents.
CommNet has notably been extended for abstractive summarization in natural language processing~\cite{celikyilmaz2018deep}.
VAIN~\cite{hoshen2017vain} can be seen as a CommNet variant with an attention mechanism.
It can effectively model high-order interactions with linear complexity in the number of agents while preserving the structure of the problem.
Finally, and particularly relevant to this work, BiCNet~\cite{peng2017multiagent} proposed to use a bidirectional recurrent network to connect individual agent’s policy and value networks in two ``information flow'' directions.
We note that it has shown outstanding performance in StarCraft micromanagement tasks.

The above methods are based on pre-defined communication architectures, which restricts the flexibility of communication.
Moreover, they all require constant communication between agents.
In real-world applications, constant communication can be costly, since receiving a large amount of information requires high bandwidth and high computational complexity.
To tackle these difficulties, some recent work has started to focus on dynamic communication architectures.
For example, G2ANet~\cite{liu2020multi} first proposed the use of hard- and soft-attention mechanisms to indicate whether communication between two agents should happen by predicting the importance of such communication; this work then relies on a graph neural network to explicitly learn communications among agents.
ATOC~\cite{jiang2018learning} added an attention unit to both determine when communication is needed and how to combine received information.
Similarly, IC3Net~\cite{singh2018learning} uses a gating mechanism to decide when to communicate so that the model can work in multi-agent cooperative, competitive and mixed settings.
TarMAC~\cite{das2019tarmac} allows agents to communicate for multiple rounds before taking actions in the environment and use a simple signature-based soft attention mechanism to decide what messages to send and whom to address them to.
Through learning the importance of each agent's partially observed information, SchedNet~\cite{kim2019learning} achieves communication scheduling, i.e., deciding which agents should be entitled to broadcast their encoded messages.
These methods can allow agents to learn a number of dynamic communication protocols, but as a result of their desire to minimize communication, can perform poorer in problems that may require (or can offer) global communication among agents, such as the ones considered in this work.

Finally, in addition to establishing effective communication protocols, we note that a number of studies also focuses on understanding agents’ communication content. 
Mordatch and Abbeel~\cite{mordatch2018emergence} investigated how basic language among agents is generated and the meaning of these abstract discrete symbols.
Kottur et al.~\cite{kottur2017natural} showed that the language could be made more human-like by placing certain restrictions in a discrete setting with two agents.


\section{Background}


\subsection{Decentralized Partially-Observable Markov Decision Process}

This paper considers fully cooperative multi-agent tasks with $n$ agents. These tasks can be described as a Decentralized partially-observable Markov decision process (Dec-POMDP).
For $n$ agents, the  Dec-POMDP is defined by ($S,\left\{A_{i}\right\}, T, R,\left\{\Omega_{i}\right\}, O, \gamma$), where $i\in\{1,2,3 \ldots . n\}$ denotes agent $i$, $s \in S$ is the global state of the environment, $a_{i} \in A_{i}$ denotes the action of agent $i$ and $\mathrm{a}=\left\{a_{1}, a_{2}, \ldots a_{n}\right\}$ is the joint action of agents, $T$ is a set of transition probabilities between states with $T\left(s, a, s^{\prime}\right)=P\left(s^{\prime} \mid s, a\right)$, $R$ is the reward function, $o_{i} \in \Omega_{i}$ is the partial observation of agent $i$, $O$ is a set of observation probabilities with $O\left(s^{\prime}, a, o\right)=P\left(o \mid s^{\prime}, a\right)$, $\gamma$ is a discount factor.
At each time step $t$, each agent $i$ first receives a partial observation $o_{i}$ based on the observation probabilities $O$.
Then, agent $i$ selects an action $a_{i}$ according to its individual policy $p_{i}$.
The joint action $\mathrm{a}=\left\{a_{1}, a_{2}, \ldots a_{n}\right\}$ is executed, updating the environment to the next state $\mathrm{s}^{\prime}$ based on the transition function $T$, and returning a global reward $r_{t}=R(s, a)$ for the whole team.
The goal of the agents is to maximize the discounted return $J=\sum_{t=0}^{l} \gamma^{t} r_{t}$,  with $l$ the episode time horizon.

\begin{figure*}[t]
  \centering
  \setlength{\belowcaptionskip}{-0.3cm}
  \includegraphics[width=0.8\linewidth]{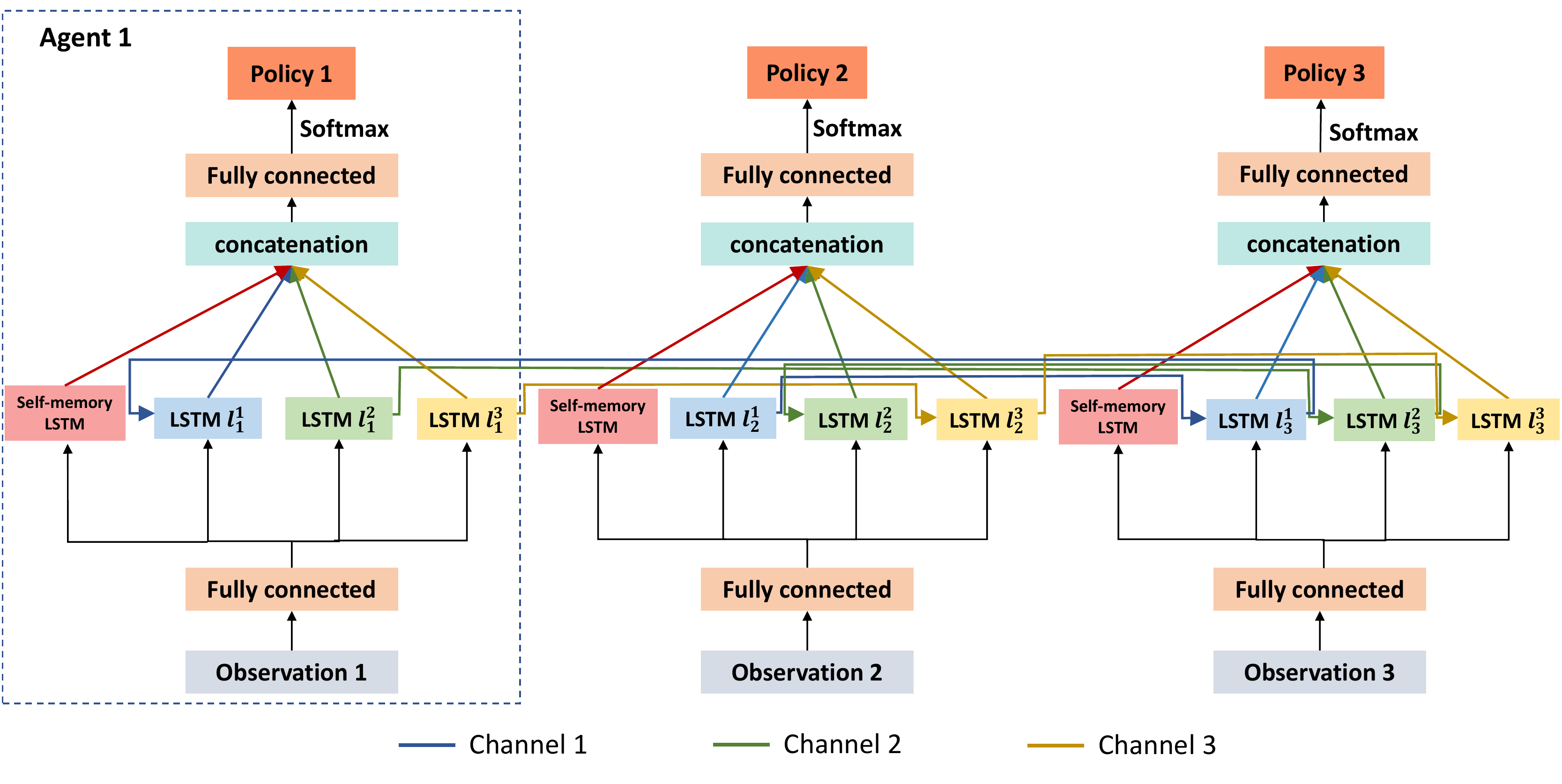}
  \vspace{-0.55cm}
  \caption{Structure of the policy network in FCMNet, for an example task with three agents and therefore three parallel communication channels (arrows between agents in the middle of the network).
  Communications flow in the direction of these arrows, connecting the agent's LSTM units to form three parallel, one-directional communication channels.
  There, the output hidden state and cell state of each LSTM unit act as messages.
  These messages are sequentially transmitted between agents along each channel; all communications happen in the same time step (i.e., multi-hop communications).
  We also consider a separate critic network, which has the same structure but a single state value estimation output for each agent.
  We use parameter sharing among agents for both the actor and critic networks, but no parameter sharing among these two networks.
  }
  \label{fig:FCMNet}
  \Description{Policy network structure of FCMNet in an example task with three agents, and therefore three parallel communication channels (arrows between agents in the middle of the network).
  Communications flow in the direction of the arrows, connecting the agent's LSTM units to form three parallel, one-directional communication channels.
  Specifically, the output hidden state and cell state of each LSTM unit act as messages.
  These messages are sequentially transmitted between agents along each channel; all communications happen in the same time step (i.e., multi-hop communications).
  We also consider a separate critic network, which has the same structure but a single state value estimation output for each agent.
  We use parameter sharing among agents for both the actor and critic networks, but no parameter sharing among these two networks.}
\end{figure*}


\subsection{Proximal Policy Optimization Algorithms}

Proximal policy optimization (PPO)~\cite{schulman2017proximal} is a policy gradient method with an actor-critic structure.
The most crucial difference between PPO and standard policy gradient methods is that, where standard policy gradient methods usually perform one gradient update per data sample, PPO adds clipped probability ratios to the optimisation objective of the actor, thus avoiding destructively large policy updates and enabling multiple epochs of minibatch updates.
PPO adjusts the weights $\theta$ of the actor network to maximize the objective
\begin{equation*}
L^{CLIP}(\theta)=\widehat{\mathbb{E}}_{t}\left[\min \left(r_{t}(\theta) \hat{A}_{t}, \operatorname{clip}\left(r_{t}(\theta), 1-\epsilon, 1+\epsilon\right) \hat{A}_{t}\right)\right],
\end{equation*}
\noindent where $r_{t}(\theta)$ denotes the clipped probability ratio  $r_{t}(\theta)=\frac{\pi_{\theta}\left(a_{t} \mid s_{t}\right)}{\pi_{\theta_{old}}\left(a_{t} \mid s_{t}\right)}$, with $\pi_{\theta}$ and $\pi_{\theta_{old}}$ the new and old policies of the agent, respectively, $\epsilon$ a clipping hyperparameter (often set to $\epsilon = 0.2$), and $\hat{A}_{t}$ the truncated version of the generalized advantage function.
By taking the minimum of the clipped and unclipped objective, PPO effectively bounds the rate of change of the agent's policy.
In addition, PPO also adds a policy entropy term to the final objective function of the actor, which improves exploration by discouraging premature convergence to suboptimal, deterministic policies.

In this paper, we rely on the multi-agent version of PPO to train agents under the framework of centralized training and decentralized execution (CTDE) .
We adapt both the network structure of actor and critic to integrate our communication protocol.

\section{Full Communication Memory Net}

In this section, we detail the neural structure of FCMNet.
We first describe our communication channel, which is based on directional recurrent neural networks (RNNs).
We then present our self-memory unit, which is used to allow agents to maintain an internal state and aggregate their own observations over time.
We note that FCMNet is applicable to general multi-agent problems that satisfy the following conditions:
\vspace{-0.1cm}
\begin{itemize}
\item All agents in the environment are able to communicate with each other at each time step.
\item Agents are able to have multiple rounds of communication within one time step (multi-hop capabilities).
\end{itemize}

\subsection{Full Communication Protocol}

In general, FCMNet is based on an actor-critic structure under the CTDE framework where both the actor and critic of each agent only transmit information in the communication layer.
The remaining layers are separated and all their parameters are shared among all agents in the team.
Through the CTDE framework, the critic is able to get additional information during training, which leads to faster convergence.
Weight sharing also helps speed up the learning process by relying on the bulk of experience from all agents.
In addition, weight sharing between agents also leads to a form of invariance in the agents' policy.
That is, FCMNet can avoid learning inefficient policies where only one agent is active, and the other agents do not contribute to the team.

\vfill

The global multi-hop communication protocol is the key of FCMNet.
It is based on parallel directional RNNs connecting the agents along different sequences, thus forming parallel communication channels.
Specifically, and as illustrated in Figure~\ref{fig:FCMNet}, for a task with $n$ agents, we assign $n$ Long Short-Term Memory (LSTM) units to each agent, and connect the LSTM units in a non-repetitive sequence to form $n$ communication channels, where each agent only has a single LSTM unit in each communication channel.
The LSTM units set of agent $i \in\{1,2,3 \ldots n\}$ is denoted as $L_{i}=\left(l_{i}^{1}, l_{i}^{2}, l_{i}^{3} \ldots l_{i}^{n}\right)$.
Communication channel $j \in\{1,2,3 \ldots n\}$ is composed of a set of LSTM units $C_{j}=(l_{1}^{j}, l_{2}^{j}, l_{3}^{j} \ldots l_{n}^{j})$, where $l_{i}^{j}$ represents the LSTM unit of agent $i$ in communication channel $j$.
The parameters of the LSTM units of different communication channels are different, but the parameters of the LSTM units in the same communication channel are shared among all agents.
The LSTM unit can be seen as an information extractor/encoder, and in FCMNet, the hidden state and cell state output by each LSTM unit are the messages transmitted between agents.
The hidden state and cell state of the LSTM unit of the previous agent will be used as the initial hidden state and cell state of the LSTM unit of the following agent to form a one-directional message flow communication channel that connects all agents in a fixed sequence (for each communication chain).
In each communication channel, the input message (namely the initial hidden state and cell state) of the first LSTM unit is always a zero vector, while the output of the last LSTM unit is unused.

\vfill

The communication channels constructed by these sequences of LSTM units offer an effective way of encoding high-level information\hfill~in\hfill~latent\hfill~space\hfill~into\hfill~a\hfill~fixed-length\hfill~message\hfill~that\hfill~is\hfill~sequentially
\newpage \noindent
transmitted to agents.
That is, FCMNet agents learn to sequentially integrate their own observation ''on top'' of information from other agents upstream in the communication channel, allowing each one to receive and contribute to ongoing flows of information throughout the team.
Since this operation happens before the action selection, agents' decisions are indeed conditioned on their own observation as well as on messages exchanged within the team.
Furthermore, since these communication channels are differentiable, they are trained by using gradients from all agents’ policy or critic losses.
That is, such differentiable CL allows FCMNet agents to explicitly inform each other of exactly how received messages guided them towards better/worse actions, thus improving each others’ action selection through backpropagation during centralized training.
Communication channels will be trained to capture more useful information that lead to reduced agents' losses, thereby leading to enhanced-cooperation policies.
This key capability to identify and share relevant observation information and propagate gradients among all agents are most likely reasons why we are able to report improved performance in highly-cooperative tasks, which can only be attained when agents consensually act as a single coherent unit.

For each agent to receive extracted information from all other agents, we introduce a simple fixed connection topology of LSTM units in communication channels.
In the communication channel $i$, the LSTM unit of agent $i$ is always in the last position of the RNN, and other agents are connected in front of it in a fixed and non-repetitive sequence (in practice, we let channel $i$th follow the sequence $1,\dots,i-1,i+1,\dots,n,\textbf{i}$).
Therefore, the message received by agent $i$ on the $i$th communication channel includes information processed by all agents before it.
We experimentally found that the sequence of other agents in front of the last position does not have a significant impact on the final performance of FCMNet.
The final output of the communication protocol of agent $i$ is the concatenation of the output of its LSTM unit set $L_{i}=(l_{i}^{1}, l_{i}^{2}, l_{i}^{3} \ldots l_{i}^{n})$, since the messages received by the agent in all communication channels all contains useful information, even though some of these messages contain information extracted by a portion of the team only.

Note that the final output of the communication protocol comes from multiple communication channels, and the received messages have been processed multiple times by other agents.
Therefore, even if one received message contains inaccurate/noisy information, this message will likely be diluted in the set of the other, more accurate messages, often still allowing efficient decision-making.
In summary, the global and multi-hop communication features of FCMNet endow agents with some natural ability to resist interference, by introducing a form of messaging redundancy over the multiple parallel communication channels.
In addition, the proposed, simple topology allows not only the length of each communication channel to be equal to the number of agents, but the number of parallel communication channels to grow linearly with the team size.


\subsection{Self-Memory Unit}

In addition to the recurrent units forming the parallel communication channels, each agent also has an additional LSTM unit for itself, called the \textit{self-memory unit}.
Its input hidden state and cell state are its output hidden state and cell state at the previous time step, and its input is the agent's current observation.
Through the self-memory unit, agents are able to integrate past information across time.
The idea of the self-memory unit originates from Deep Recurrent Q-Learning~\cite{hausknecht2015deep}. 
However, our model does not directly input the output of the self-memory unit to the next neural layer, but concatenates it with the output of the communication protocol.

We have selected the structure described above for our self-memory unit, since it performed best through a large set of control experiments.
There, we varied self-memory unit's input, the input state of its hidden state and cell state, and the combination method of its output and that of the communication protocol.
Specifically, we tried version of this unit where we fed as input 1) the agent’s current observation, 2) the concatenation of all received messages, and 3) the concatenation of the agent’s current output hidden states and cell states.
We further tried versions of the this unit where the hidden state and cell state input were 1) the output hidden state and cell state of self-memory unit at the previous time step, and 2) simply two zero vectors.
Finally, we both tried to concatenate the output of this unit to the output of the communication protocol (winning strategy), as well as directly feeding the output of this unit to the next neural layer (i.e., in cases where the outputs of the communication protocol were fed into the self-memory unit).


\section{Experiments}

In this section, we first benchmark FCMNet\footnote{The full code is available at \url{https://github.com/marmotlab/FCMNet}} against a set of value-based and communication-based baseline algorithms on a standardized partially-observable StarCraft II micromanagement environment with shared reward, called SMAC\cite{samvelyan2019starcraft}.
We then further investigate the robustness of FCMNet under four different realistic communication disturbances in a collaborative multi-agent pathfinding task with individual rewards.

For all tasks, both the critic and actor of FCMNet have three hidden layers.
The second layer is the communication layer, and the number of hidden units of each LSTM unit is 64.
We set the PPO clipping parameter to $0.2$, discount factor to $0.99$, and use the Adam optimizer.
The learning rate and number of updates per epoch vary by task (see code for these details).
We train on $16$ parallel environments, each running $512$ steps in the SMAC tasks and $2048$ steps in our pathfinding task before performing a training step.
We adopt the following evaluation procedure: for SMAC tasks, the training is paused after every $5000$ steps, at which point $16$ evaluation episodes with agents greedily enacting the current policy in a decentralized manner.
The percentage of those episodes in which agents defeat all enemy units within the time limit is referred as the evaluation win rate (higher is better).
For the pathfinding task, the pause interval is $100 000$ steps, and the number of evaluation episodes remains $16$. The average episode length of these evaluation episodes is referred to as the evaluation episode length (where lower is better).


\subsection{Performance Experiments}

\subsubsection{StarCraft II Micromanagement with Shared Reward}

SMAC is built based on the strategy game StarCraft II. 
However, unlike a regular full game of StarCraft II that requires advanced actions such as gathering resources, planning buildings, this environment only simulates a battle between two platoons of units to evaluate how well independent agents are able to cooperate to solve complex skirmish tasks.
In each scenario, one army is controlled by a reinforcement learning algorithm, in which each unit is an independent learning agent.
The other army is controlled by the built-in, non-learned, heuristics game AI.

SMAC has been widely used as a standard environment in multi-agent experiments.
Some recent work has improved algorithm performance by changing the output of the environment~\cite{yu2021surprising}.
In order to compare our approach fairly, we have kept the default setting of SMAC in this work.
We consider the following standard SMAC tasks in our experiments, in increasing order of difficulty: $2m\_vs\_1z$, $3m$, $2c\_vs\_64zg$, $3s\_vs\_3z$, $3s\_vs\_4z$, $10m\_vs\_11m$, $5m\_vs\_6m$.
These are all fully cooperative and homogeneous multi-agent tasks, but differ in the units attributed to the two platoons.
The overall goal is to eliminate enemy units, namely, maximize the win rate.
An episode terminates when all enemy units have died or when the episode reaches the pre-defined time limit.
A game is counted as a win only if all enemy units are eliminated. 
Partial observability is achieved by introducing a circular, unit field-of-view area, which allows the agent only to receive information about two armies and terrain features within this field-of-view.
The global state -- that agents are unable to receive during execution -- contains information about all units on the task.
The action space of an agent is discrete, it consists of moving in four directions, stopping, attacking a certain enemy unit, and no-op that can only be executed by a dead agent. 
Enemy units can only be attacked when they are within the agent’s circular shooting area.
This facilitates the decentralization of the problem and forces agents to explore cooperative behaviors.
Agents share a team reward in all SMAC tasks, which is simply the total damage dealt to enemy units at each time step.
Additionally, agents receive a $+10$ reward after killing an enemy unit, and $+200$ reward after winning the skirmish.


\subsubsection{Result}

We compare FCMNet with 6 standard baselines in the field, namely CommNet~\cite{sukhbaatar2016learning}, G2ANet~\cite{liu2020multi}, SchedNet~\cite{kim2019learning}, IQL~\cite{tan1993multi}, VDN~\cite{sunehag2018value} and QMIX~\cite{rashid2018qmix} on the 7 different SMAC tasks considered.
CommNet, G2ANet and SchedNet are all differentiable CL methods, closer to our approach.
CommNet uses the average value of hidden states from all agent modules as a communication message and allows multiple rounds of communication in one timestep.
G2AN uses hard-attention and soft-attention to indicate whether there is communication between two agents and the importance of the communication.
SchedNet selects $k$ out of $n$ agents to broadcast their encoded messages in each timestep by learning to estimate the importance of each agent’s partial observation to the team.
$k$ is manually predefined and can be changed for different tasks.
In our experiments, we always set $k = n-1$, that is, only one agent is unable to broadcast its messages in each timestep.
We let agents in FCMNet, Commnet, G2ANet, and SchedNet communicate $n$, 3, 1, and 1 time(s) per timesteps, respectively.
IQL, VDN and QMIX all belong to the class of Q-learning methods, with no learned communications.
IQL simply treats a multi-agent problem as a collection of multiple single-agent problems sharing the same environment; this is the most vanilla baseline considered.
The individual agent is trained by DQN~\cite{mnih2015human}, without any explicit interaction among agents.
VDN and QMIX improve the performance of standard Q-learning by different value decomposition methods.
VDN uses the sum of individual value functions to represents the team's value function.
QMIX deploys a neural network whose weights are derived from the global state to combine individual value functions into the team's value function in a non-linear fashion.
Value decomposition methods, despite not endowing agents with communication abilities, often perform better at fully-cooperative tasks, as they allow us to explicitly address the credit assignment problem, which in turn provides agents with a more accurate learning signal that often lead to subtle cooperative maneuvers at the team level (i.e., closer to joint decision-making).

\begin{figure}[t]
  \vspace{0.1cm}
  \centering
  \setlength{\belowcaptionskip}{-0.3cm}
  \includegraphics[width=\linewidth]{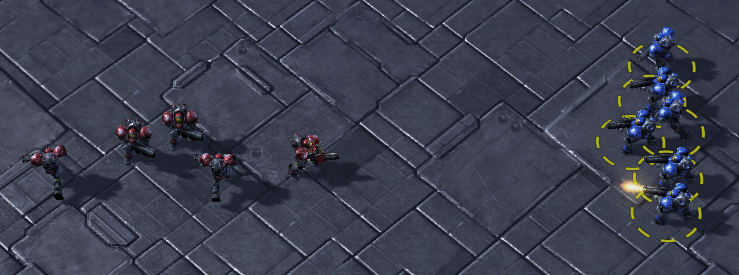}
  \vspace{-0.65cm}
  \caption{$5m\_vs\_6m$ task in SMAC (hardest task considered in this work). RL agents control the 5 Marines on the left, outnumbered in their skirmish against the 6 Marines on the right, control by the game's built-in, non-learning-based AI.}
  \label{fig:starcraft}
  \Description{$5m\_vs\_6m$ task in SMAC (hardest task considered in this work). RL agents control the 5 Marines on the left, outnumbered in their skirmish against the 6 Marines on the right, control by the game's built-in, non-learning-based AI.}
\end{figure}

\begin{table*}[t]
    \caption{Mean evaluation win rate and standard deviation on all the SMAC tasks considered for different algorithms, using 10M training timesteps. The score of the best-performing algorithm(s) for each task is highlighted in bold.}
    \vspace{-0.25cm}
    \label{tab:evaluation}
    \begin{tabular}{ cc|ccc|c|cc}\toprule
    & \textbf{FCMNet} & \textbf{CommNet} & \textbf{G2ANet} & \textbf{SchedNet} & \textbf{IQL} & \textbf{VDN} & \textbf{QMIX} \\ \midrule
    $2m\_vs\_1z$ & \textbf{100.0(0.0)} & 93.8(10.8) & \textbf{100.0(0.0)} & \textbf{100.0(0.0)} & 98.4(2.7) & 96.9(5.4) & 96.9(3.1) \\
    $3m$ & \textbf{100.0(0.0)} & 90.6(3.1) & \textbf{100.0(0.0)} & \textbf{100.0(0.0)} & \textbf{100.0(0.0)} & \textbf{100.0(0.0)} & \textbf{100.0(0.0)} \\
    $2c\_vs\_64zg$ & \textbf{100.0(0.0)} & 79.7(14.9) & 89.1(5.2) & 95.3(2.7) & 28.1(10.4) & 81.3(7.7) & 93.8(6.3) \\
    $3s\_vs\_3z$ & \textbf{100.0(0.0)} & 0.0(0.0) & 0.0(0.0) & \textbf{100.0(0.0)} & \textbf{100.0(0.0)} & \textbf{100.0(0.0)} & 98.4(2.7) \\
    $3s\_vs\_4z$ & 92.2(5.2) & 0.0(0.0) & 0.0(0.0) & 89.1(5.2) & 96.9(3.1) & \textbf{98.4(2.7)} & \textbf{98.4(2.7)} \\
    $10m\_vs\_11m$ & 71.9(10.4) & 0.0(0.0) & 1.6(2.7) & 0.7(0.3) & 15.6(12.9) & 78.1(7.0) & \textbf{85.9(5.2)} \\
    $5m\_vs\_6m$ & 40.6(21.9) & 0.0(0.0) &  0.0(0.0) & 0.0(0.0) & 28.1(3.1) & 18.8(0.0) & \textbf{59.4(3.1)} \\\bottomrule
    \end{tabular}
\end{table*}

\begin{figure*}[t]
\vspace{-0.4cm}
\captionsetup[subfigure]{aboveskip=-1pt} 
\centering
\includegraphics[width=0.8\textwidth]{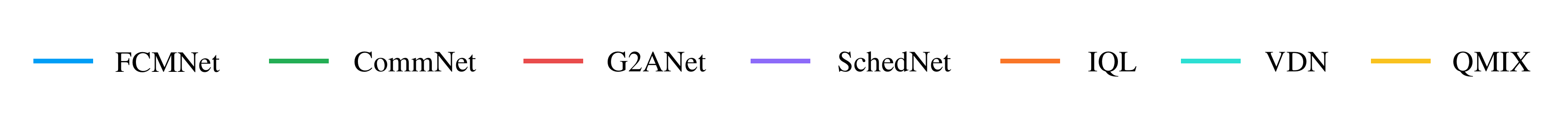}\\[-0.4cm]
\begin{subfigure}[b]{0.25\textwidth}
    \includegraphics[width=\textwidth]{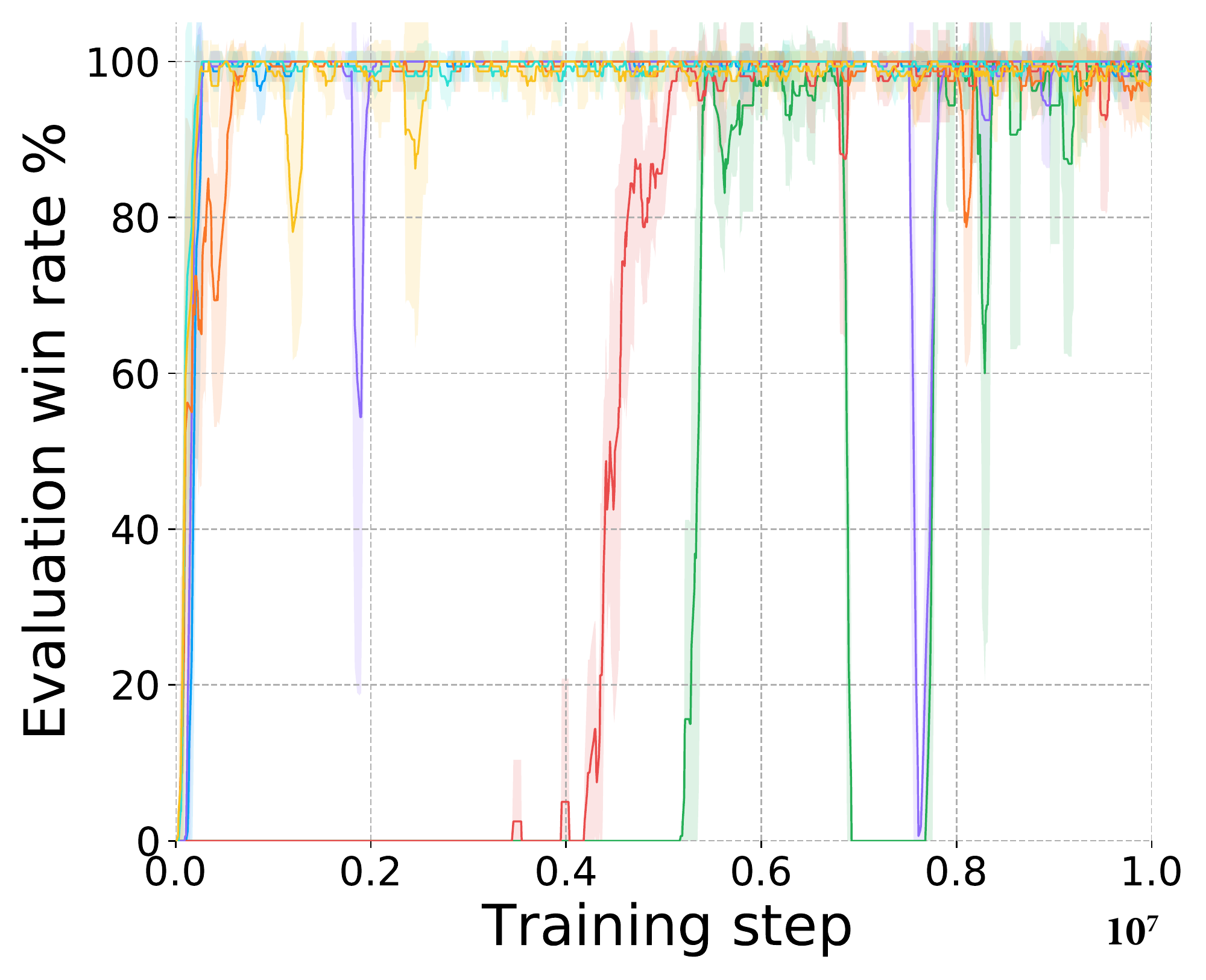}
    \caption{2m\_vs\_1z}
    \label{fig:2m1z}
\end{subfigure}
\begin{subfigure}[b]{0.24\textwidth}
    \includegraphics[width=\textwidth]{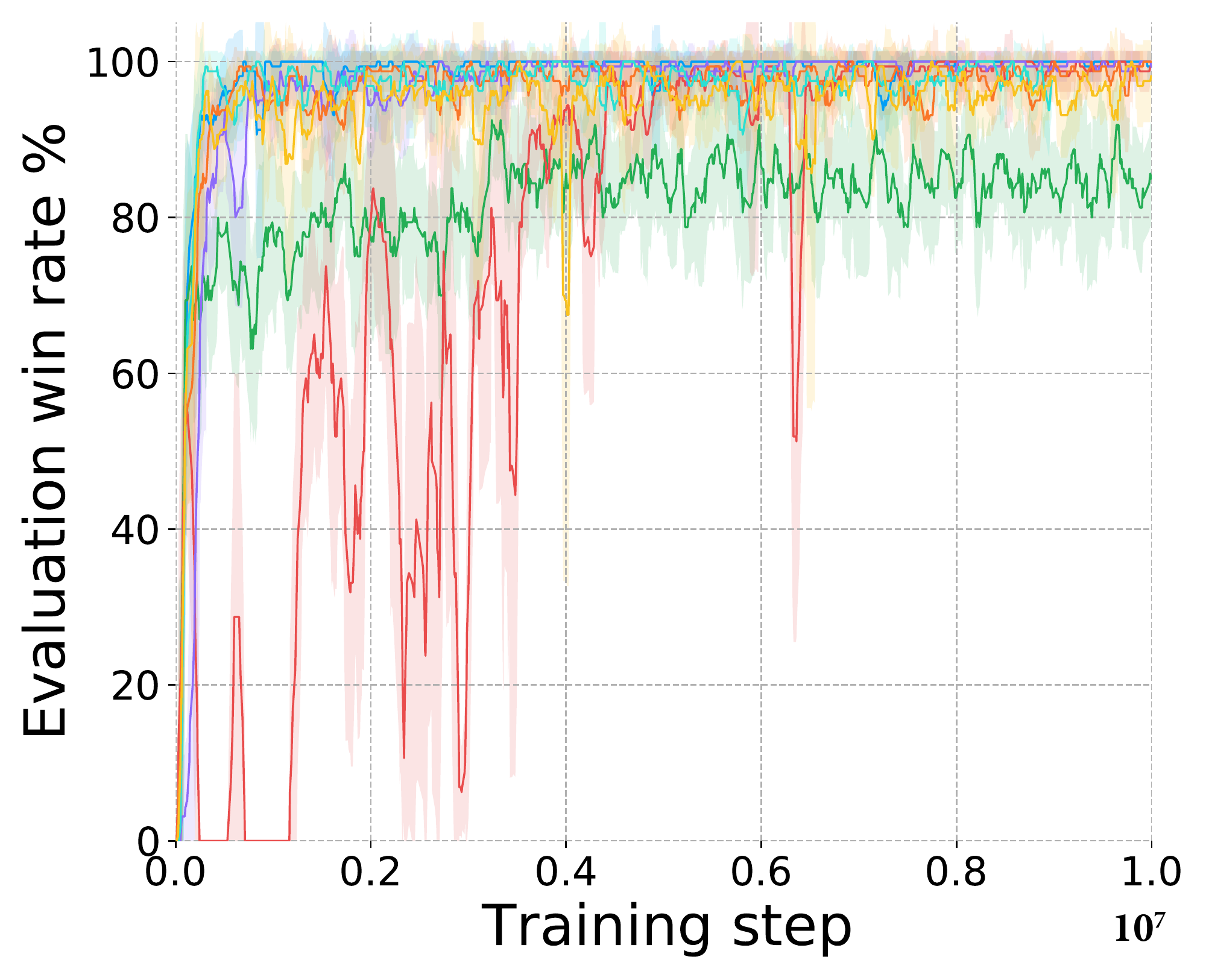}
    \caption{3m}
    \label{fig:3m}
\end{subfigure}
\begin{subfigure}[b]{0.24\textwidth}
    \includegraphics[width=\textwidth]{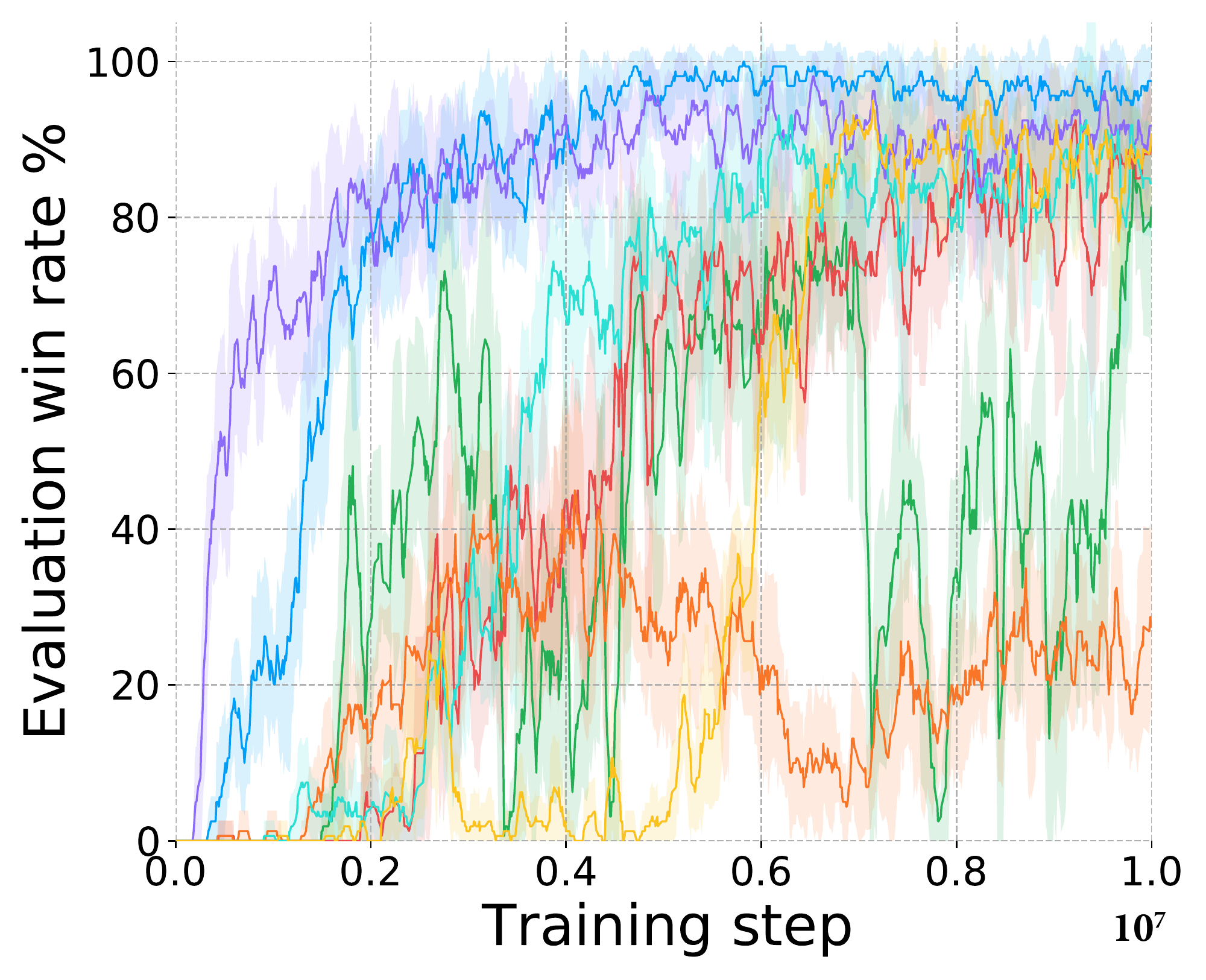}
    \caption{2c\_vs\_64zg}
    \label{fig:2c64zg}
\end{subfigure} \\[0.1cm]
\begin{subfigure}[b]{0.24\textwidth}
    \includegraphics[width=\textwidth]{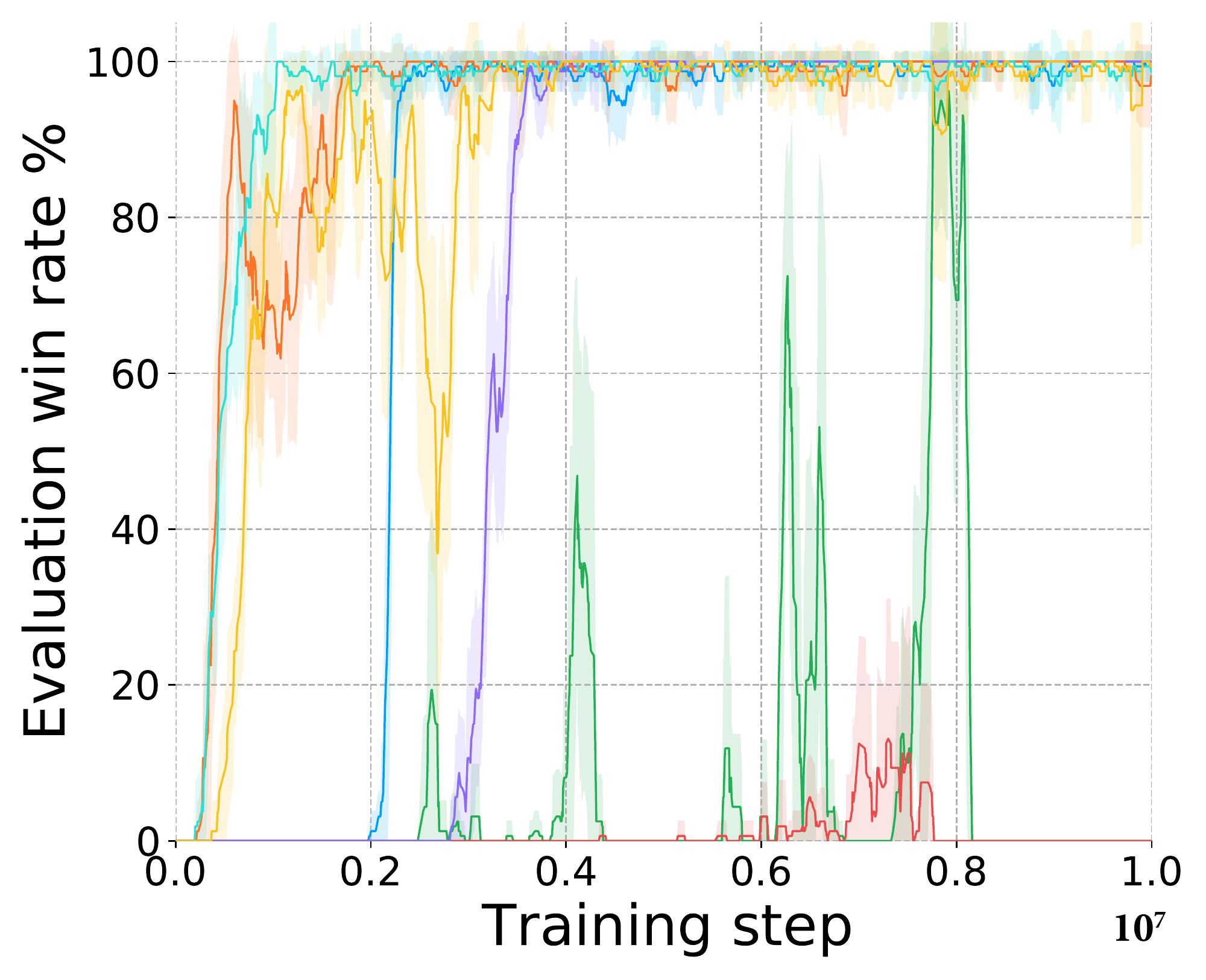}
    \caption{3s\_vs\_3z}
    \label{fig:3s3z}
\end{subfigure}
\begin{subfigure}[b]{0.24\textwidth}
    \includegraphics[width=\textwidth]{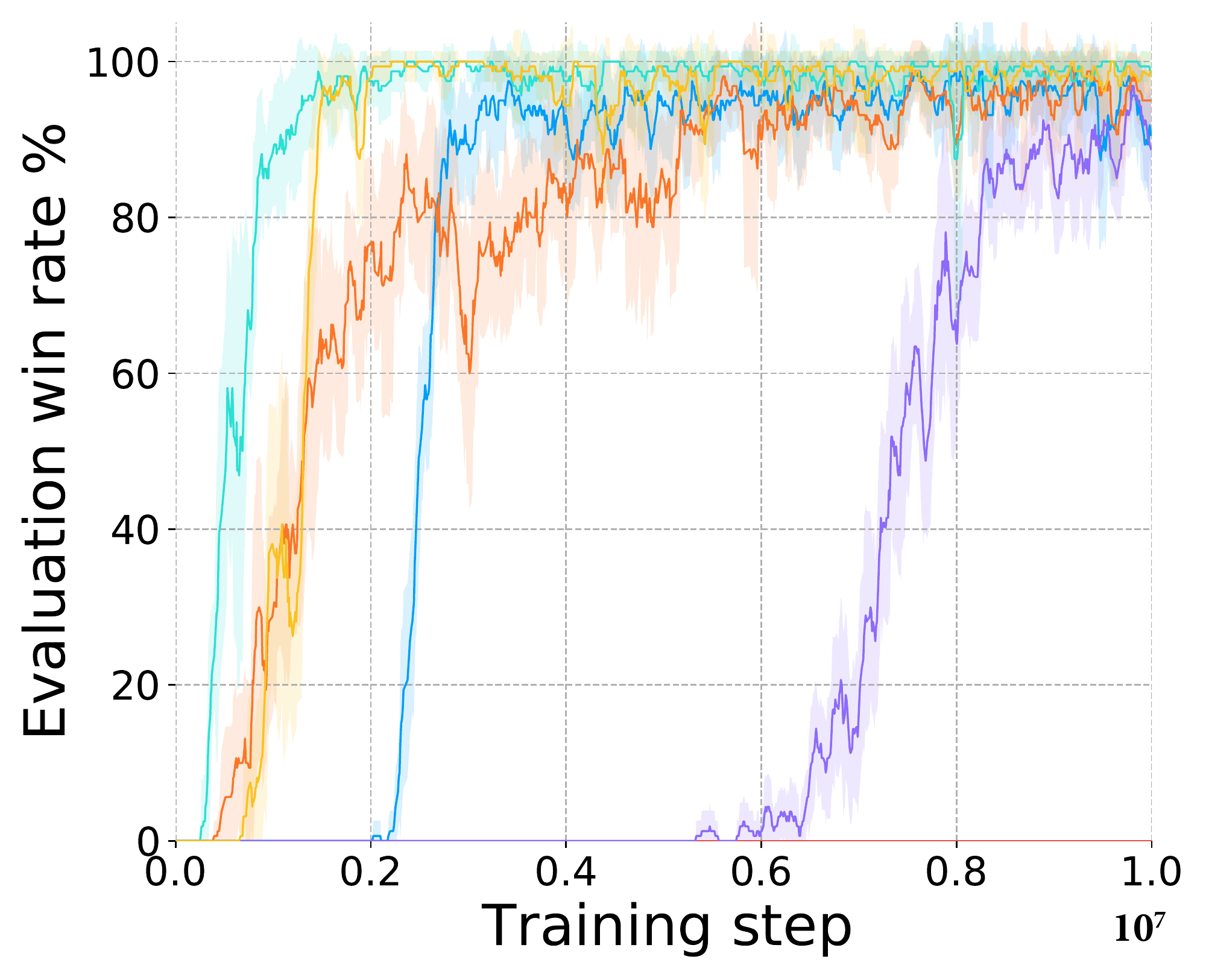}
    \caption{3s\_vs\_4z}
    \label{fig:3s4z}
\end{subfigure}
\begin{subfigure}[b]{0.24\textwidth}
    \includegraphics[width=\textwidth]{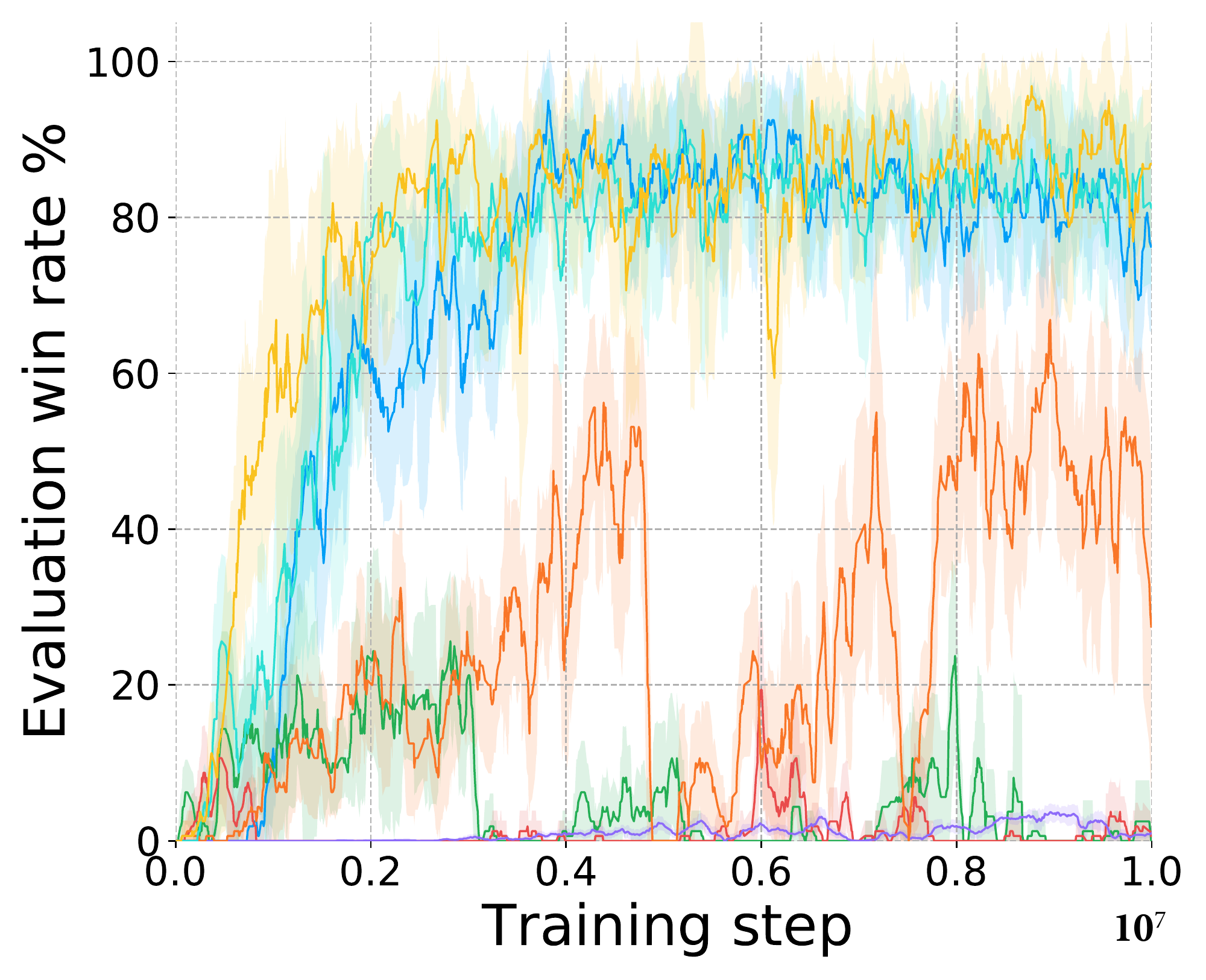}
    \caption{10m\_vs\_11m}
    \label{fig:10m11m}
\end{subfigure}
\begin{subfigure}[b]{0.24\textwidth}
    \includegraphics[width=\textwidth]{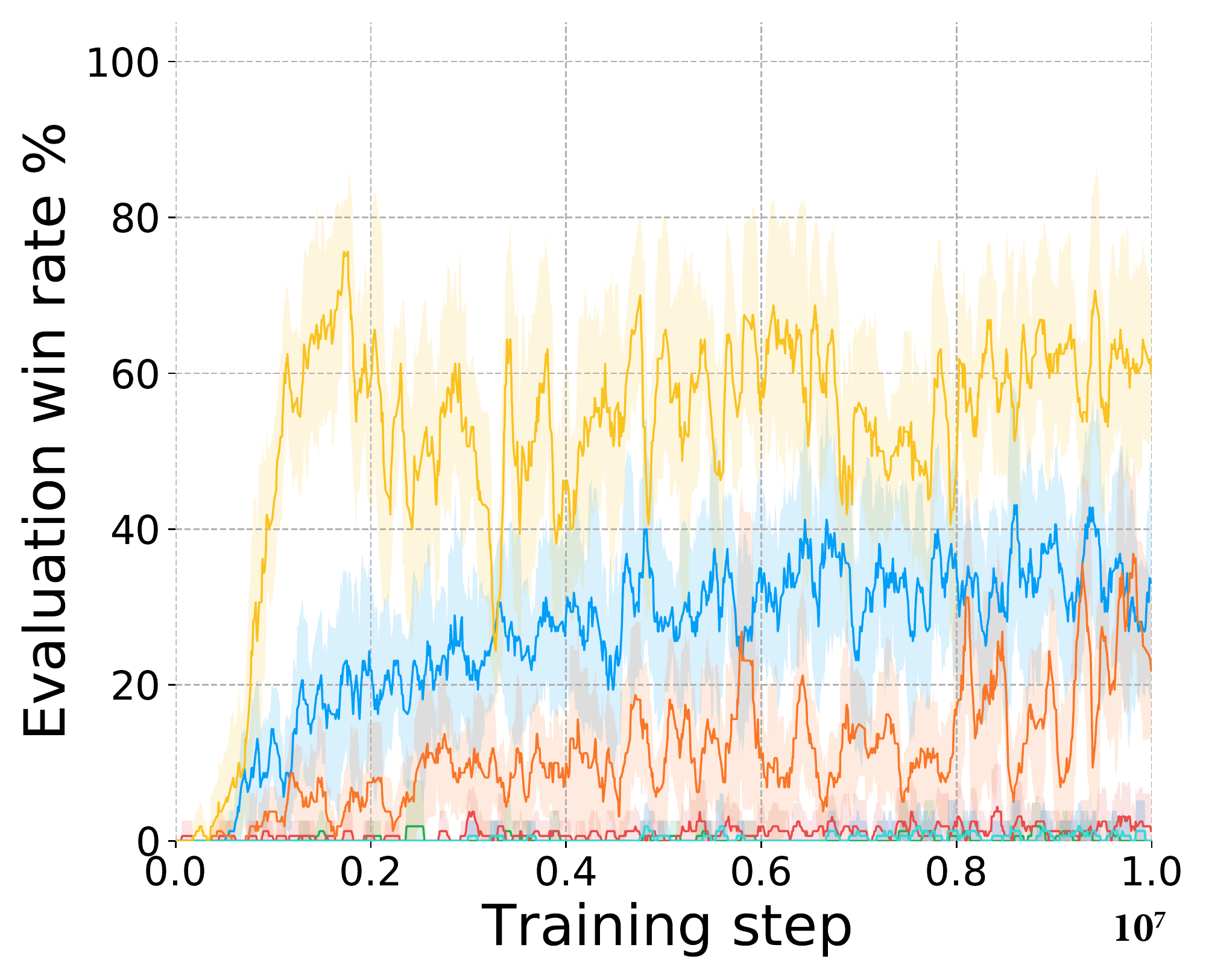}
    \caption{5m\_vs\_6m}
    \label{fig:5m6m}
\end{subfigure}
\setlength{\belowcaptionskip}{-0.5cm}
\vspace{-0.3cm}
\caption{Training curves of different algorithms on SMAC tasks, showing the average win rate.
The confidence interval (shaded area) shows one standard deviation over 160 evaluation episodes.
CommNet, G2ANet, SchedNet and FCMNet are all differentiable communication learning methods, while IQL, VDN, and QMIX are based on Q-learning with no learned communications. In particular, VDN and QMIX rely on value decomposition, and as expected, exhibit improved performance on harder tasks.}
\label{evaluation_figure}
\vspace{0.3cm}
\end{figure*}

Our first, general observation, as shown in Table~\ref{tab:evaluation} and Figure~\ref{evaluation_figure}, is that FCMNet is able to outperform all communication-based baselines on all tasks, sometimes quite significantly (e.g., for harder tasks).
In tasks with lower difficulty such as $2m\_vs\_1z$ and $3m$, G2ANet, SchedNet and FCMNet all reach a win rate of 100\% easily, but training curves indicate that FCMNet converges faster and exhibits more stable/consistent performance.
Moreover, FCMNet beats the communication-based baselines with significant margins in the $3s\_vs\_3z$, $3s\_vs\_4z$, $10m\_vs\_11m$ and $5m\_vs\_6m$ tasks, where FCMNet is still able to perform rather effectively, while the final win rate of CommNet and G2ANet falls close to 0\%.
FCMNet performs comparably to the value decomposition methods in all 7 tasks, which is rather impressive considering the difficuly of the last task in particular ($5m\_vs\_6m$).
In $2m\_vs\_1z$ and $3s\_vs\_3z$ tasks, FCMNet, VDN and QMIX quickly reach a high win rate, but the win rate of VDN and QMIX fluctuates slightly after the training curve converges.
Therefore, the final mean evaluation win rate of FCMNet is $100\%$, while the final mean evaluation win rate of VDN and QMIX is slightly lower.
The significant performance difference between FCMNet, VDN and QMIX can be seen in $2c\_vs\_64zg$ and $5m\_vs\_6m$ tasks.
In the $2c\_vs\_64zg$ task, the final mean evaluation win rate of FCMNet is 18.7\% and 6.2\% higher than VDN and QMIX, respectively.
In the $5m\_vs\_6m$ task, the win rate of FCMNet is 21.8\% higher than VDN, but is defeated by QMIX with a gap of 18.8\%.
We believe that the performance degradation of FCMNet is due to its inability to explicitly address the credit assignment problem, which seems critical for the hardest SMAC task considered.
As a result, individual agents are unable to effectively determine their contribution to the team reward, and truly find their place to offer a coherent winning strategy.
In addition, we also noticed that, although IQL gets an acceptable win rate in most tasks, its training curves are highly unstable due to the non-stationarity of the environment, which arises due to other agents constantly changing their behavior during training.

We visually examined the learned behaviors of the policies to interpret the superior performance of FCMNet.
A selection of these videos is available in the supplemental material.
Agents trained by FCMNet seem to have learned basic skirmish skills in easier tasks, namely quickly finding the closest enemy and approaching it to attack until it is dead, after which agents will immediately attack the next closest enemy.
Agents also perform more advanced maneuvers depending on the task.
On the $2m\_vs\_1z$ task, because the enemy's attack range is limited, both agents learn to inflict damage at range, while avoiding the enemy's attacks.
Furthermore, we often observe an advanced cooperative behavior on this task, whereby one of the agents will lead the enemy away from the other agent by successive back-and-forth movements (to attract and keep the enemy's focus), so that second agent can safely attack the enemy (i.e., collaborative ``\textit{kiting}'').
On the $3m$ task, agents and enemy units are equal in number and strength.
Therefore, agents learn to first ``\textit{focus-fire}'' a specific enemy, and then kill other enemies in sequence, a well-known, optimal strategy for symmetric skirmishes.
Finally, on the $3s\_vs\_3z$ task, when all enemies are alive at the beginning of an episode, agents will first divide themselves into two groups: two agents overpowering a single enemy, while the last agent ``kites'' the remaining two enemies.
Once the outnumbered enemy has been killed, the two agents reunite with their teammate, quickly overpowering the remaining two enemies.

FCMNet, CommNet, G2ANet, and SchedNet all explicitly learn communication mechanisms.
However, our results indicate that FCMNet consistently outperforms these differentiable CL baselines.
We believe that there are two key points to FCMNet's success.
First, CommNet uses the average value of hidden states from all agent modules as a communication message.
This averaging operation implicitly treats messages from different agents equally.
Differently, FCMNet uses independent directional RNNs to encode information, thus implicitly allowing incoming messages to be processed with different weights before concatenation.
We believe this flexibility is important, as messages from different information flows might contain different aspects/take on the current sequential reasoning of the team, and therefore should be processed and used differently into the final decision of each agent.

Second, contrary to the full communication network of FCMNet, G2ANet and SchedNet both implement a dynamic communication mechanism, that is, each agents cannot communicate with all other agents at every timestep.
Although this mechanism can reduce communication overload/redundancy, we believe that full communication might be beneficial (or even needed) in the SMAC tasks, as it allows agents to reach the type of near-joint/-consensual decision-making required for platoon-based skirmishes.


\subsection{Robustness Experiments}

\subsubsection{Multi-Agent Pathfinding with Individual Rewards}

To further illustrate the robustness of FCMNet, we consider the simple partially-observable, cooperative multi-agent pathfinding task introduced in~\cite{freed2020communication} (\textit{Hidden-Goal Path-Finding}), with a team of $n=5$ agents.
In this task, each agent has a unique target location it needs to reach as soon as possible, and whose position may change randomly at every timestep.
An episode terminates when the distances between each agent and its respective target are all less than a threshold ($0.01$ in practice), or when the episode length reaches $1024$ steps.
Each agent's observation contains the location of other agents' targets but \textbf{not its own target}.
As a result, effective communication is needed to solve this problem.
Each agent can apply five discrete actions, namely, applying a unit force in each of the four cardinal directions, and no force.
Different from SMAC which considers global rewards, the rewards in this task are assigned individually to each agent, based on the distance between it and its target and an constant time step penalty.


\subsubsection{Binarized Messages}

The communication protocol in FCMNet is differentiable and optimized through backpropagation, where gradients flow among agents through the communication channel.
This gives agents richer feedback, thus easing the discovery of effective cooperative policies.
Therefore, our algorithm tends to converge faster and to higher-quality policies, compared to traditional RL techniques that treat communication as actions, and observe their effect from later rewards (i.e., reinforced CL approaches).

However, most modern communication technologies rely on discrete communication channels, for which continuous communication cannot be directly applied (i.e., bitwise/digital communications).
To further investigate the practicality of FCMNet under such realistic constraints, we convert the original continuous real-valued message of FCMNet (i.e., the hidden state and cell state of each agent's LSTM units) with a length of 128, into a binary message with a length of 20 by adding a stacked autoencoder and a binarization step into the actor's communication channels.

\begin{figure}[t]
  \centering
  \setlength{\belowcaptionskip}{-0.55cm}
  \includegraphics[width=\linewidth]{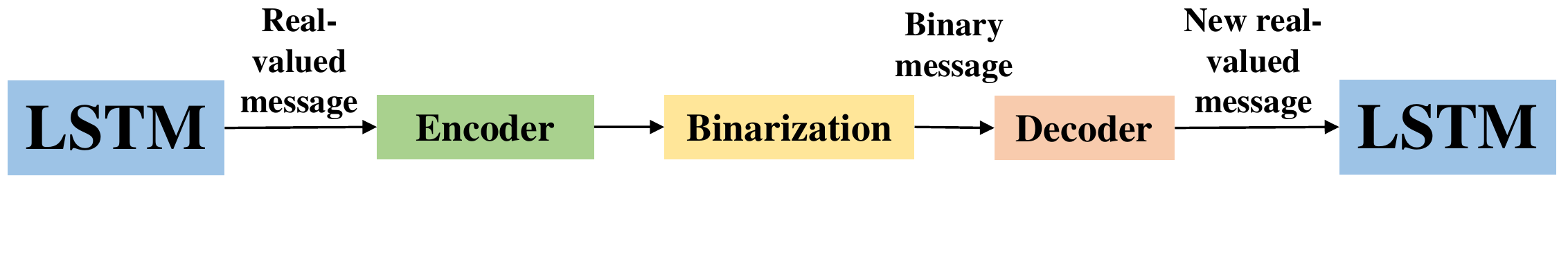}
  \vspace{-1.2cm}
  \caption{Binarization/De-binarization process, introduced between each two successive LSTMs (i.e., between communicating agents) in the FCMNet variant with binary messages.}
  \label{fig:binary}
  \Description{Binarization/De-binarization process, introduced between each two successive LSTMs (i.e., between communicating agents) in the FCMNet variant with binary messages.}
  \vspace{0.3cm}
\end{figure}

The neural network structure between two LSTM units of FCMNet with binarized messages is presented in Figure~\ref{fig:binary}, the weights of the autoencoder are shared among agents in one communication channel, but can be different among communication channels.

The binarization process we use is inspired by~\cite{courbariaux2015binaryconnect,williams1992simple,toderici2015variable}, and consists of two steps.
The first step, i.e., the \textit{encoder}, generates the required length of outputs in the continuous interval $[-1,1]$, achieved by a fully-connected layer with a tanh activation function.
The second step, i.e., the binarization, produces discrete data in the set $\{-1,1\}$ based on the continuous output of the encoder:
\begin{equation*}
b(x)=x+\epsilon \in\{-1,1\},
\end{equation*}
where $\epsilon \in\{1-x,-x-1\}$  is a random variable distributed according to $P(\epsilon=1-x)=\frac{1+x}{2}$ and $P(\epsilon=-x-1)=\frac{1-x}{2}$. 
Therefore, the complete binarization process is:
\begin{equation*}
B(x)=b\left(\tanh \left(\omega^{t-1} x+b^{t-1}\right)\right),
\end{equation*}
where $\omega^{t-1}$ and $b^{t-1}$ are the weights and bias of the fully-connected layer with tanh activation function.

\begin{figure}[t]
  \centering
  \includegraphics[width=0.4\linewidth]{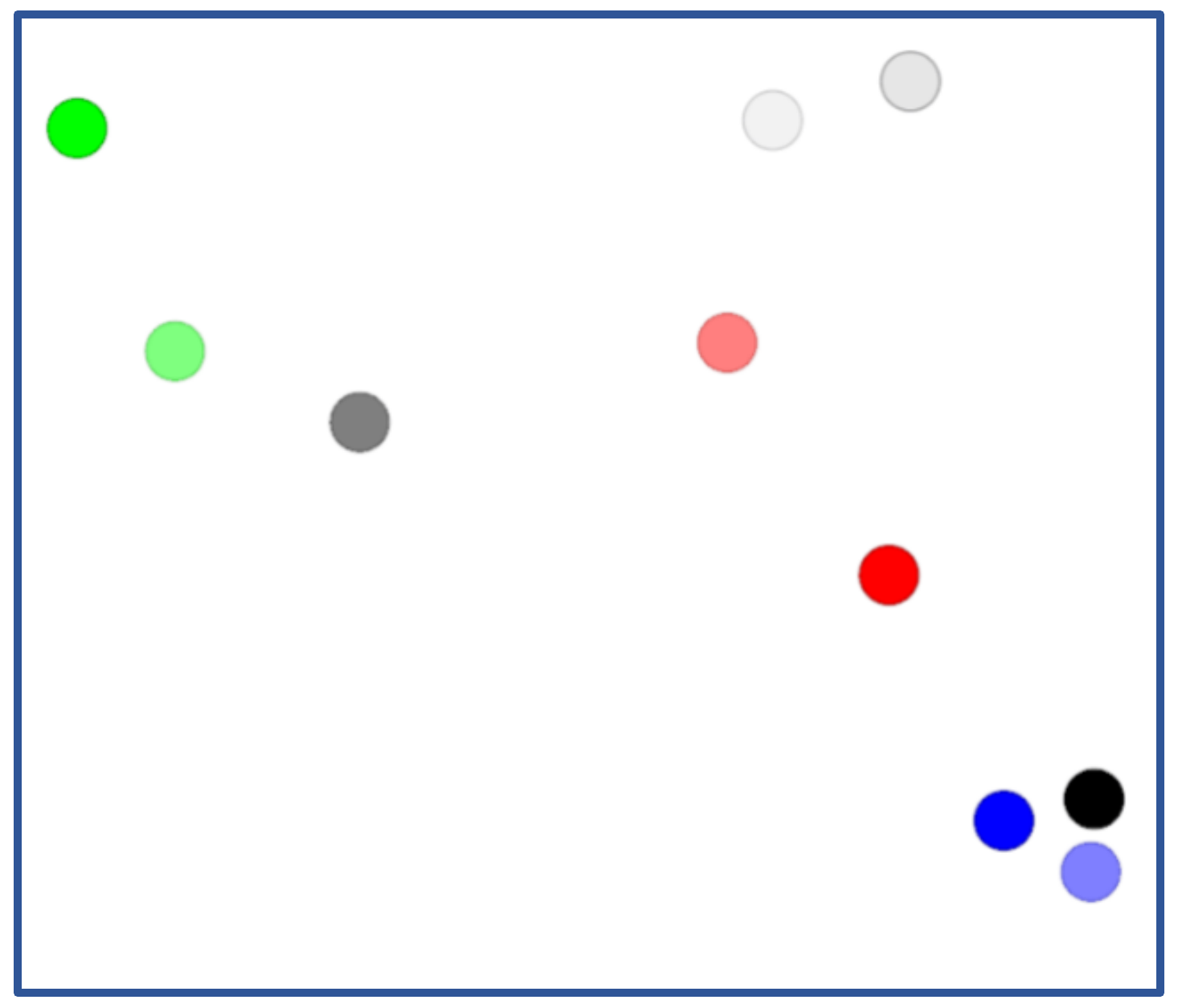}
  \vspace{-0.25cm}
  \caption{Multi-agent partially-observable pathfinding task. Each agent (saturated colored dot) must reach its goal (same-colored, less saturated dot) as fast as possible. Goals change locations at random time steps. Agents get access to each others' position, and to the position of all but their goal, which they must obtain via communications.}
  \label{fig:Path finding}
  \Description{Multi-agent partially-observable pathfinding task. Each agent (saturated colored dot) must reach its goal (same-colored, less saturated dot) as fast as possible. Goals change locations at random time steps. Agents get access to each others' position, and to the position of all but their goal, which they must obtain via communications.}
  \vspace{-0.3cm}
\end{figure}

\begin{figure*}[t]
\captionsetup[subfigure]{aboveskip=1pt} 
\centering
\setlength{\belowcaptionskip}{-0.4cm}
\begin{subfigure}[b]{0.28\textwidth}
    \includegraphics[width=\textwidth]{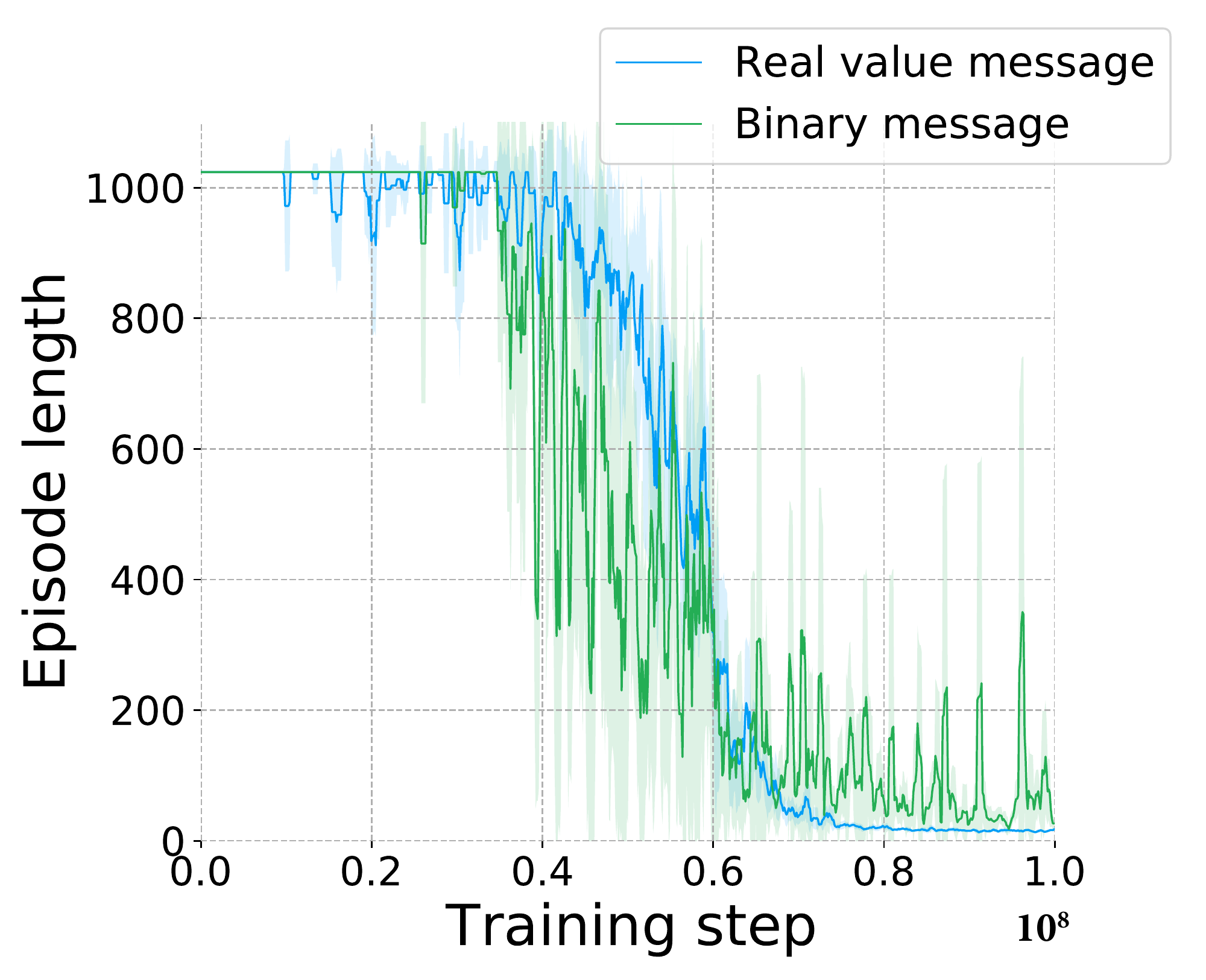}
    \caption{Binarized messages}
    \label{binary_result}
\end{subfigure} \hfill
\begin{subfigure}[b]{0.28\textwidth}
    \includegraphics[width=\textwidth]{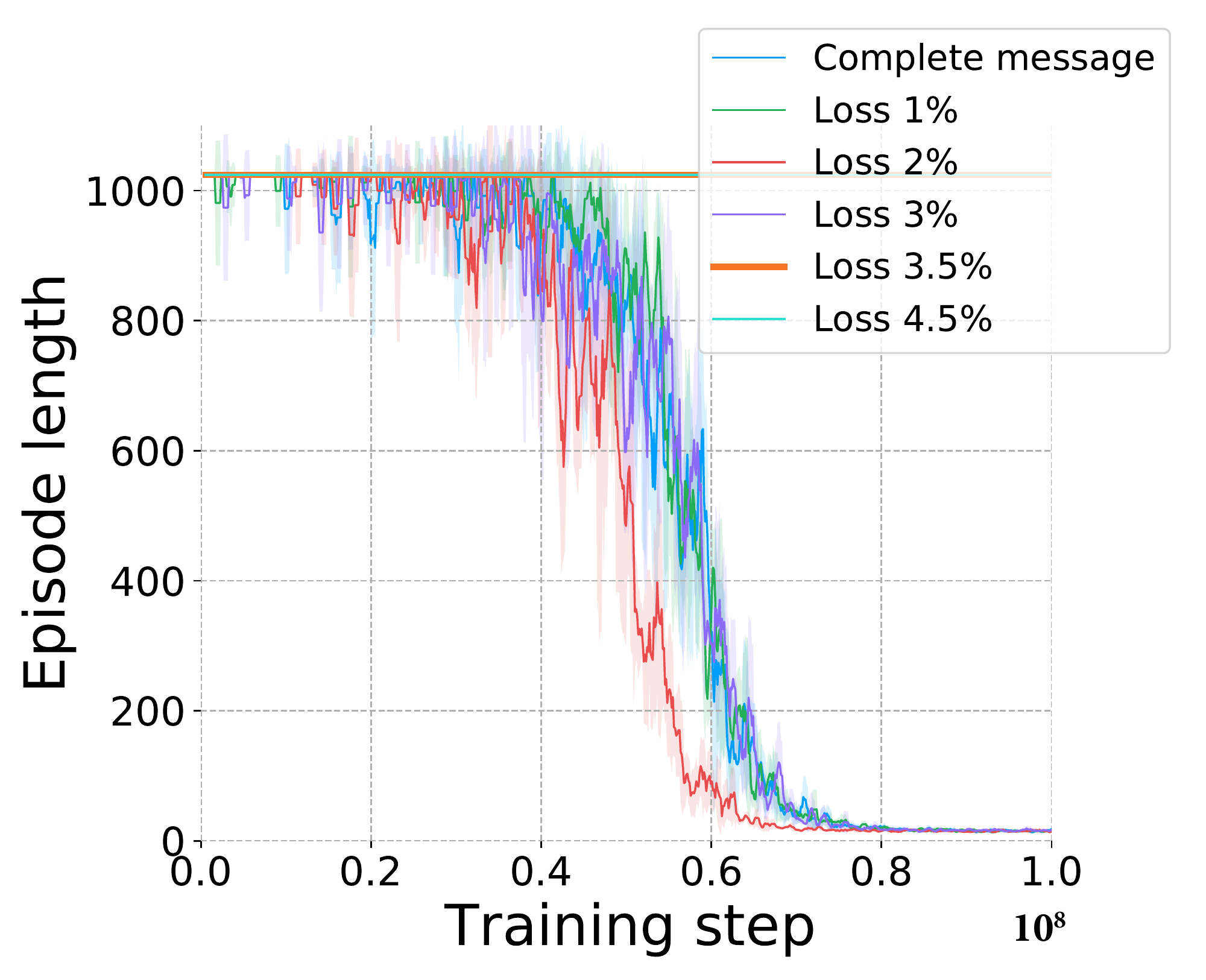}
    \caption{Lost messages}
    \label{fig:loss}
\end{subfigure} \hfill
\begin{subfigure}[b]{0.28\textwidth}
    \includegraphics[width=\textwidth]{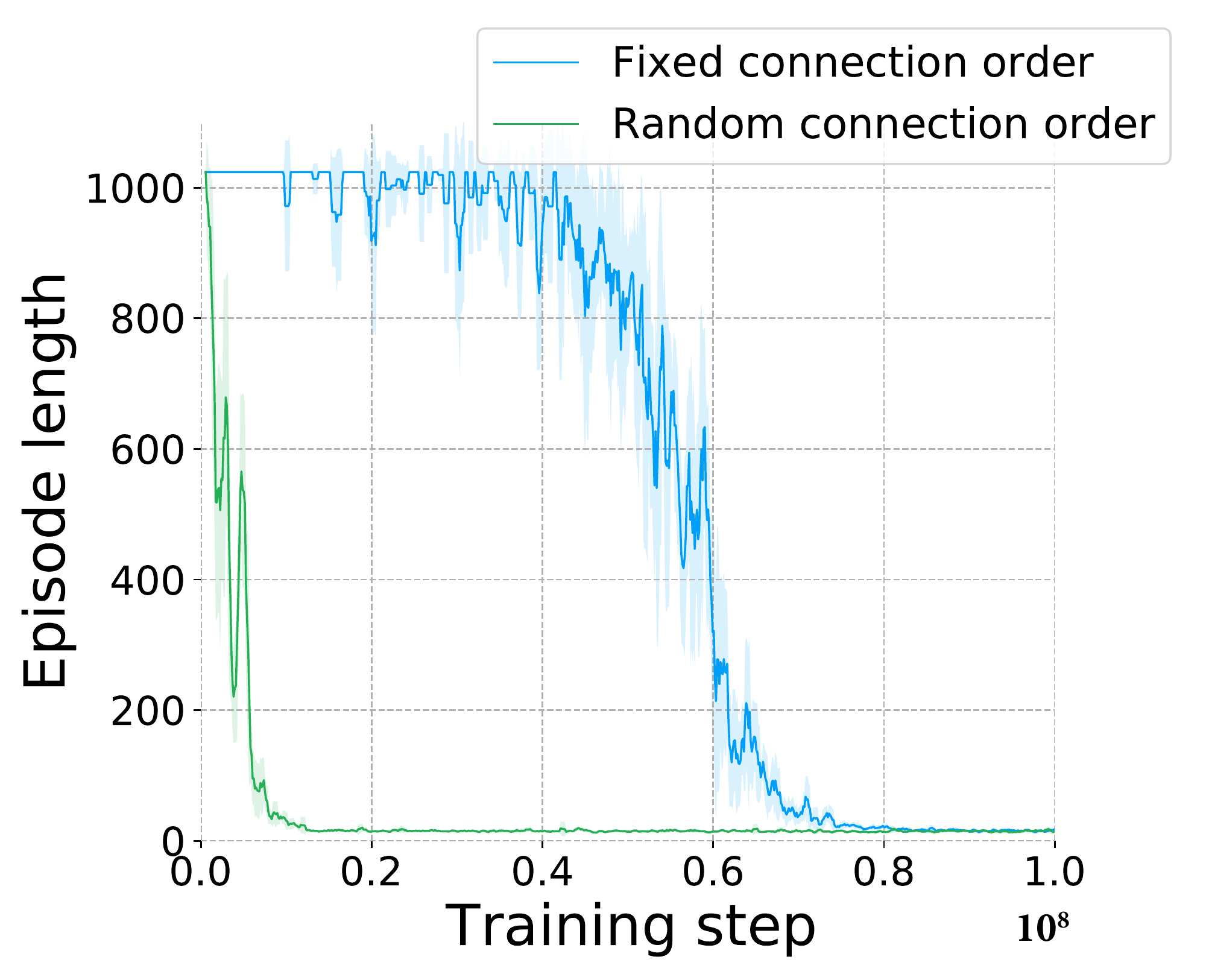}
    \caption{Random communication order}
    \label{fig:random}
\end{subfigure}
\vspace{-0.1cm}
\caption{Training curves for our robustness investigation.
The confidence interval (shaded area) shows one standard deviation over 80 evaluation episodes.
These plot show the average number of steps to complete a pathfinding task, where shorter episodes are better.
Our results show that FCMNet still converges to the same level of performance, even under three communication disturbance: binarized messages, random message loss, and randomized communication orders at each time step.\vspace{0.3cm}}
\Description{Experiments results for robustness performance}
\end{figure*}

The process of message transmission is shown in Figure~\ref{fig:binary}, the message sender first converts a real value message into a binary message through the encoding and binarization process.
The receiver agent then inputs the binary message into a symmetrical decoder to obtain a new real value message.
The new real value message is used as the input hidden state and cell state of the next LSTM unit, following the standard FCMNet structure.

Figure~\ref{binary_result} shows the learning speed and converged episode length of FCMNet with binarized messages in the multi-agent pathfinding task, which remain similar to FCMNet with real-valued messages, while naturally handling more realistic real-world communication constraints.
However, the training of FCMNet with binarized messages is more unstable, even if the general performance is improved with training.
The reasons behind this is that the gradient propagation between agents is interrupted by the binarization processing, and a significant amount of information is lost during the conversion/deconversion process.
They both increase the difficulty of learning collaboration between agents.


\subsubsection{Lost Messages}

In wireless digital communication, there are many factors that can cause information loss, such as noise interference, information transmission delay, hardware damage, etc.
Therefore, we believe that the performance of FCMNet under random message loss is key in practical applications.
In these experiments, we assume that agents have a fixed probability of losing a complete message at every message transmission step, which then gets replaced with a zero vector of the same length.

In order to find the threshold of model collapse, we conducted multiple tests with different message loss probabilities in the multi-agent pathfinding task.
The main result are presented in Figure~\ref{fig:loss}.
These show that, when the probability of message loss is equal to or lower than 3\%, the final converged episode length and converge speed are unaffected, and remain identical to the standard FCMNet model without message loss.
However, when the probability of message loss increases above 3.5\%, FCMNet stops working entirely.
In this second case, the actor and critic losses both remain turbulent and cannot be reduced during training.
Therefore, we conclude that FCMNet can naturally resist (limited) random message loss, where the probability threshold for this task lies between 3\%-3.5\%.

We examine the learned policy in order to understand the influence of lost messages better.
The agents trained by FCMNet with message loss probability equal to or lower than 3\% can quickly reach the position of their targets regardless of whether the targets are moving randomly or not.
However, agents trained by FCMNet with message loss probability equal to or higher than 3.5\% can only get close to their target but cannot reach the target position accurately.
When targets change their position, the agents need to wander for a while before they can obtain information about their target's new position and approach it again.
That is, we believe that, when the frequency of lost messages is too high, agents cannot consistently extract the exact target position from the messages received at every time step with high accuracy.


\subsubsection{Random Communication Order}

In previous experiments, the structure of FCMNet was fixed, that is, agents always transmit messages in a pre-determined, fixed sequence along each communication channel.
However, we believe this fixed sequence might not be flexible enough for practical applications.
In this experiment, we further explore the performance of FCMNet when the order of message transmission changes randomly at every time step.

Figure~\ref{fig:random} shows the evaluation curve of normal FCMNet and FCMNet with random connection order.
Our results indicate that FCMNet with random connection order not only converges to the same performance level as normal FCMNet, but seems to exhibit faster convergence in the multi-agent pathfinding task.
We believe that this increased training speed might be due to the fact that, when changing the communication order randomly, the diversity of messages in the communication channel is increased and makes communications richer/more varied, thus helping agents to explore their cooperative policy space faster and more uniformly.


\vfill
\section{Conclusion}

This paper focuses on the class of problems where global communications are available but may be unreliable, for which we propose FCMNet, a new multi-agent deep reinforcement learning method that simultaneously learns a decentralized policy and a multi-hop communications protocol in a centralized setting. 
FCMNet makes efficient use of the messages from all agents by multiple directional recurrent neural networks, which enables team-level decision-making and improves global cooperation.
In our results on the Starcraft II Multi-Agent Challenge, FCMNet is shown to outperform state-of-the-art communication-based methods and achieves high-quality results, comparable to value decomposition methods.
Additionally, we further investigate the natural robustness of FCMNet in a multi-agent pathfinding task, under different realistic communication disturbances, such as message binarization, random message loss, and random communication sequences.

Future work will extend FCMNet to handle teams of heterogeneous agents, where weight sharing might not be possible or may need to be used on parts of the neural structure only, and to more general communication topologies (one-to-one, one-to-many, many-to-many, etc.).
Finally, we acknowledge that our work has been mostly simulation-based at this stage, and future work will focus on deployment of FCMNet on physical robots in cooperative tasks under real-life conditions and communication constraints.


\vfill
\begin{acks}

This work was supported by the Singapore Ministry of Education Academic Research Fund Tier 1.
We would like to thank Mehul Damani and Benjamin Freed for their feedback on earlier drafts of this paper.
We are also grateful to Benjamin Freed, Rohan James and Yizhuo Wang for very helpful research discussions.

\end{acks}


\newpage
\bibliographystyle{ACM-Reference-Format} 
\bibliography{mybibfile}


\end{document}